\documentclass[conference]{IEEEtran}
\IEEEoverridecommandlockouts

\makeatletter
\let\MYcaption\@makecaption
\makeatother

\usepackage[font=footnotesize]{subcaption}

\makeatletter
\let\@makecaption\MYcaption
\makeatother

\usepackage{bm}
\usepackage{url}
\usepackage{cite}
\usepackage{amsmath,amssymb,amsfonts}
\usepackage{algorithm}
\usepackage{algorithmic}
\usepackage{graphicx}
\usepackage{textcomp}
\usepackage{xcolor}
\def\BibTeX{{\rm B\kern-.05em{\sc i\kern-.025em b}\kern-.08em
    T\kern-.1667em\lower.7ex\hbox{E}\kern-.125emX}}
\DeclareMathOperator*{\argmin}{arg\,min}

\begin{document}

\title{No-PASt-BO: Normalized Portfolio Allocation Strategy for Bayesian Optimization\\
\thanks{This article was part of a project financed through the R\&D program of the Brazilian Electric Energy Agency (ANEEL), with number PD-00063-3045 and funded by CPFL Energy and Companhia Energetica Rio das Antas (CERAN). The authors also acknowledge the support from Delfos Intelligent Maintenance.}
}

\author{\IEEEauthorblockN{Thiago de P. Vasconcelos, Daniel A. R. M. A. de Souza, C\'{e}sar L. C. Mattos and Jo\~{a}o P. P. Gomes}
\IEEEauthorblockA{\textit{Department of Computer Science} \\
\textit{Federal University of Cear\'{a}}\\
Fortaleza-CE, Brazil \\
thiagodpvasconcelos@hotmail.com, danielramos@lia.ufc.br, cesarlincoln@dc.ufc.br, jpaulo@dc.ufc.br}
}

\maketitle

\begin{abstract}
Bayesian Optimization (BO) is a framework for black-box optimization that is especially suitable for expensive cost functions. Among the main parts of a BO algorithm, the acquisition function is of fundamental importance, since it guides the optimization algorithm by translating the uncertainty of the regression model in a utility measure for each point to be evaluated. Considering such aspect, selection and design of acquisition functions are one of the most popular research topics in BO. Since no single acquisition function was proved to have better performance in all tasks, a well-established approach consists of selecting different acquisition functions along the iterations of a BO execution. In such an approach, the GP-Hedge algorithm is a widely used option given its simplicity and good performance. Despite its success in various applications, GP-Hedge shows an undesirable characteristic of accounting on all past performance measures of each acquisition function to select the next function to be used. In this case, good or bad values obtained in an initial iteration may impact the choice of the acquisition function for the rest of the algorithm. This fact may induce a dominant behavior of an acquisition function and impact the final performance of the method. Aiming to overcome such limitation, in this work we propose a variant of GP-Hedge, named No-PASt-BO, that reduce the influence of far past evaluations. Moreover, our method presents a built-in normalization that avoids the functions in the portfolio to have similar probabilities, thus improving the exploration. The obtained results on both synthetic and real-world optimization tasks indicate that No-PASt-BO presents competitive performance and always outperforms GP-Hedge.             
\end{abstract}


\section{Introduction}

Bayesian optimization (BO, \cite{Mockus1991}) is a sequential design strategy based on the Bayesian framework for global optimization of unknown functions, i.e., possibly noisy functions without known closed-form expression and gradient information. It has been widely used in the most diverse tasks, such as selection of hyperparameters for machine learning algorithms \cite{Snoek2012,klein2016fast,kotthoff2017auto,falkner2018bohb,feurer2019hyperparameter}, control policies in robotics \cite{calandra2016bayesian,chatzilygeroudis2017black}, automated circuit design \cite{lyu2018batch,torun2018global}, etc.

The BO approach is especially useful in scenarios where the objective function is costly to evaluate. The inherent uncertainty considered by the Bayesian methodology allows for a more efficient exploration of the optimization domain with respect to the number of queried points. This is critical in applications where each individual evaluation usually involves substantial financial and/or computational effort. The uncertainty with respect to the function to be optimized usually is modelled using the Gaussian Process (GP) framework \cite{rasmussen2006}.

One of the main ingredients of the BO methodology is the so-called acquisition function, which translates the uncertainty in the task domain to a simple to evaluate function that quantifies the expected \textit{utility} of each point. Such function is then optimized to select the next point to be evaluated, i.e., the next candidate solution. However, although several acquisition functions have been proposed in the literature, there is not a single choice that is always better for any task \cite{hoffman2011portfolio}.

Hoffman \textit{et al.} \cite{hoffman2011portfolio} tackled the aforementioned issue by proposing a hierarchical hedging approach for managing an adaptive portfolio of acquisition functions based on their past performances. Similar approaches have been pursued by other authors. Shahriari \textit{et al.} \cite{shahriari2014entropy} expands the original hedge by proposing a choice criterion based on information theoretic considerations. Although in a multi-armed bandit learning context, Shen \textit{et al.} \cite{shen2015portfolio} also consider a portfolio-based strategy for balancing exploration and exploitation during the sequential decision procedure. Recently, Lyu \textit{et al.} \cite{lyu2018batch} proposed an alternative strategy that considers the multi-objective optimization of multiple acquisition functions to obtain a Pareto front from where candidate points can be sampled in a batch fashion.

Despite the above compelling recent work on the acquisition function choice problem, the resulting solutions stray from the simplicity and applicability of the original hedging strategy presented in  \cite{hoffman2011portfolio}, named GP-Hedge. Furthermore, GP-Hedge presents some undesirable properties. For instance, since it accounts for the historical performance of the individual acquisition functions to select the next candidate solution, initial discrepant values of either good or bad performance can compromise the quality of the selection strategy.

In this work we aim to propose a modified portfolio-based BO methodology that overcomes the GP-Hedge limitations while maintaining its ease of use. Our approach reduces the influence of far past evaluations to enable the recovery of initially bad performing acquisition functions and to avoid the dominance of initially good performing functions. Moreover, our proposal, named No-PASt-BO (Normalized Portfolio Allocation Strategy for Bayesian Optimization), presents a built-in normalization mechanism to avoid the functions within the portfolio to have similar probabilities of being chosen, which promotes exploration.

The new No-PASt-BO approach is empirically evaluated in the task of optimizing synthetic benchmark functions. We also consider the task of optimizing the hyperparameters of machine learning models in real world applications. The obtained results indicate that No-PASt-BO presents competitive performance and always outperform GP-Hedge.

The remaining of the paper is organized as follows. Section \ref{sec_theory} summarizes the required theoretical background; Section \ref{sec_proposed} illustrates the GP-Hedge limitations and details the proposed No-PASt-BO methodology; Section \ref{sec_experiments} presents and discuss the performed computational experiments; Section \ref{sec_conc} concludes the paper with pointers for further investigations.

\section{Theoretical Background}
\label{sec_theory}

In this section we summarize the main theoretical aspects of the BO framework, including GP basics, the role of the acquisition function and the original GP-Hedge algorithm.

\subsection{Gaussian Process Basics}
\label{sec:gp}

In a standard single dimension output regression setting, we aim to obtain a mapping $f : \mathbb{R}^{D} \rightarrow \mathbb{R}$ from a set of $N$ inputs $\bm{x}_i \in \mathbb{R}^{D}$, organized in a matrix $\bm{X} \in \mathbb{R}^{N \times D}$, to a set of $N$ correspondent outputs $f_i \in \mathbb{R}$. However, we usually only observe $\bm{y} \in \mathbb{R}^N$, a noisy version of the vector $\bm{f}$. Considering a Gaussian observation noise $\epsilon \sim \mathcal{N}(0, \sigma^2)$ we have:
\begin{equation}
\label{eq:reg}
y_i = f_i + \epsilon, \text{ where }  f_i = f(\bm{x}_i), \quad 1 \leq i \leq N.
\end{equation}

In the GP modeling framework we choose a multivariate Gaussian prior for the \textit{latent} (non-observed) vector $\bm{f}$. If a zero mean prior is chosen, we get \cite{rasmussen2006}:
\begin{align}
\nonumber
p(\bm{f}|\bm{X}) & = \mathcal{N}( \bm{f} | \bm{0}, \bm{K}_f ), \\
\nonumber
p(\bm{y}|\bm{f},\bm{X}) & = \mathcal{N}( \bm{y} | \bm{f}, \sigma^2 \bm{I}) \mathcal{N}( \bm{f} | \bm{0}, \bm{K}_f ), \\
\label{EQ:GP}
p(\bm{y}|\bm{X}) & = \mathcal{N}( \bm{y} | \bm{0}, \bm{K}_f + \sigma^2 \bm{I}), 
\end{align}
where in Eq. \eqref{EQ:GP} we were able to analytically integrate out $\bm{f}$. The elements of the covariance matrix $\bm{K}_f \in \mathbb{R}^{N \times N}$ are calculated by $[\bm{K}_f]_{ij} = k(\bm{x}_i, \bm{x}_j), \forall i,j \in \{1,\cdots,N\}$, where $k(\cdot, \cdot)$ is the so-called covariance (or \textit{kernel}) function. 

The kernel hyperparameters are usually optimized following the gradients of the log-marginal likelihood, i.e., the logarithm of Eq. \eqref{EQ:GP}, also called the \textit{evidence} of the model.

Given a new input $\bm{x}_* \in \mathbb{R}^D$, the posterior predictive distribution of the related output $f_* \in \mathbb{R}$ is calculated analytically using standard Gaussian distribution conditioning properties:
\begin{align}
\label{EQ:GP_inference}
p(f_*|\bm{y},\bm{X},\bm{x}_*) & = \mathcal{N}\left( f_* \middle| \mu_*, \sigma_*^2 \right), \\
\nonumber
\mu_* & = \bm{k}_{*f} (\bm{K}_f + \sigma_y^2\bm{I})^{-1} \bm{y}, \\
\nonumber
\sigma_*^2 & = K_* - \bm{k}_{*f} (\bm{K}_f + \sigma_y^2\bm{I})^{-1} \bm{k}_{f*},
\end{align}
where $\bm{k}_{f*} = [k(\bm{x}_*,\bm{x}_1), \cdots, k(\bm{x}_*,\bm{x}_N)]^\top \in \mathbb{R}^N$, $\bm{k}_{*f} = \bm{k}_{f*}^\top$ and $K_* = k(\bm{x}_*,\bm{x}_*) \in \mathbb{R}$. Importantly, each prediction is a fully defined distribution, instead of a point estimate, which reflects the inherent uncertainty of the regression problem.

\subsection{The Bayesian Optimization Framework}

BO is a general framework for black-box optimization. Mathematically, the problem consists in finding a global minimizer (or maximizer) of an unknown loss function \cite{brochu2010tutorial}
\begin{equation} \label{eq:BO}
    \bm{x}^* = \argmin_{\bm{x} \in \mathcal{X}}{l(\bm{x})},
\end{equation}
where $l(\cdot): \mathcal{X} \rightarrow \mathbb{R}$ denotes the loss function (or objective, when maximizing), $\mathcal{X}$ is the search space, usually given by a compact subset of $\mathbb{R}^D$, and $\bm{x}^* \in \mathbb{R}^D$ denotes the optimal solution. The loss function is usually assumed to be either hard to compute or have no simple closed form, but it can be evaluated at an arbitrary query point $\bm{x}$. Similar to Eq. \eqref{eq:reg}, the BO framework also considers cases in which we do not observe $l(\cdot)$ directly, but rather noisy observations.

BO solves the problem in Eq. \eqref{eq:BO} by sequentially querying the loss function as we keep the best-so-far candidate solution $\bm{x}^+$. In doing so, at iteration $t$, we select a new location $\bm{x}_{t+1}$ at which the loss function $l(\bm{x}_{t+1})$ is evaluated. As the method iterates, the querying points and its corresponding loss values $\mathcal{D}_t = \{\bm{x}_i, l(\bm{x}_i)\}_{i=1}^t$ are used for modeling the loss function. When a stopping criterion has been achieved, we return $\bm{x}^+$ as an approximation to the actual minimizer $\bm{x}^*$. A fundamental aspect is how to provide location guesses along iterations. For that, a probabilistic model is necessary, since $l(\cdot)$ is unknown. 

The most common probabilistic approach used in the BO framework is the GP model, summarized in Section \ref{sec:gp}. The GP model is able to quantify the uncertainty with respect the function $l(\cdot)$, especially at locations at which the function was not yet evaluated and knowledge is scarce.

The uncertainty about the loss is used to select the most promising candidate point for evaluation. It is achieved by an easy-to-compute \textit{acquisition function}. Typically, such acquisition functions are model-derived and are used to trade-off exploration and exploitation; in the sense that exploration means investigate non-explored areas (with high uncertainty), and exploitation refers to considering regions where the model prediction is high. The goal is to maximize the acquisition function, and it can be achieved because such functions are assumed to be cheap to evaluate, usually with gradient information available. 

We refer the reader to the comprehensive survey in \cite{shahriari2015taking} for more details and challenges in the general BO framework.

\subsection{Acquisition Functions}

Several acquisition functions have been proposed in the literature. Each proposal aims at measuring the quality of the candidates to be queried following a specific strategy. Next we detail three of the most used acquisition functions in practice.

\subsubsection{Probability of improvement (PI)}

The PI function, firstly proposed in \cite{kushner1964new}, focus on choosing the domain point with the highest probability of being lower than $\mu^- = \min_i \mu(\bm{x}_i)$, where $\mu(\bm{x}_i)$ indicates the GP predicted mean at the input $\bm{x}_i$. This formulation focus on exploitation \cite{hoffman2011portfolio}, which can be balanced with the trade-off hyperparameter $\xi \geq 0$ as follows:
\begin{equation}
\label{PI}
    \text{PI}(\bm{x}) = P(f(\bm{x}) \leq \mu^- - \xi) = \Phi\left(\frac{\mu^- - \xi - \mu(\bm{x})}{\sigma(\bm{x})}\right),
\end{equation}
where $\Phi$ is the CDF of a standard Gaussian distribution.

\subsubsection{Expected improvement (EI)} The EI function, introduced in \cite{Mockus1978}, considers the probability of an evaluation being lower than the current best known evaluation, but it also takes into account the magnitude of the improvement. Let $\mu^- = \min_i \mu(\bm{x}_i)$, the EI function is zero if $\sigma(\bm{x}) = 0$, otherwise it is given by
\begin{align}
\label{EI}
\text{EI}(\bm{x}) & = \tau(\bm{x})\Phi\left(\frac{\tau(\bm{x})}{\sigma(\bm{x})}\right) + \sigma(\bm{x})\phi\left(\frac{\tau(\bm{x})}{\sigma(\bm{x})}\right) \\
\nonumber
\text{where } \tau(\bm{x}) & = \mu^- - \xi - \mu(\bm{x})
\end{align}
In Eq. \eqref{EI}, $\Phi$ and $\phi$ are respectively the CDF and the PDF of a standard Gaussian distribution.

\subsubsection{GP - Lower confidence bound (LCB)} In \cite{cox1997lcb} it is introduced the ``Sequential Design for Optimization'' (SDO), which selects a point to evaluate based on the posterior mean and variance, minimizing $\mu(\bm{x}) - \kappa \sigma(\bm{x})$. In the original paper, $\kappa$ is a hyperparameter, but no clues are given in how to select it. In \cite{srinivas2010gaussian}, the SDO algorithm is revisited and distinct approaches to select a value for $\kappa$ are discussed. Thus, the so-called GP-LCB formulation is presented below:
\begin{equation}
\label{GP-LCB}
\text{GP-LCB}(\bm{x}) = \mu(\bm{x}) - \sqrt{\nu \beta_t} \sigma(\bm{x}),
\end{equation}
where we have considered $\kappa = \sqrt{\nu \beta_t}$, $\beta_t = 2\log(t^{D/2+2} \pi^2/3 \delta)$ varies with the sequential iteration $t$ and $\nu, \delta > 0$ \cite{brochu2010tutorial}.

\subsection{GP-Hedge}

As previously mentioned, there is not a single acquisition function which is the best choice for all possible optimization tasks. Therefore, a strategy which can choose among a set of acquisition functions may be a good direction to handle this issue. The GP-Hedge, introduced in \cite{hoffman2011portfolio}, follows such approach.

In the GP-Hedge framework, a set of predefined acquisition functions is considered, comprising a \textit{portfolio}. Each function nominates a candidate for the next point $\bm{x}$ of the domain to be evaluated. The candidates are then selected with a probability proportional to how good the posterior mean of the previous points the corresponding acquisition function has suggested before.

The aforementioned approach follows the \textit{hedge} strategy. According to \cite{auer1995gambling}, such method consists in choosing the action $j$ among $J$ options with probability $p_j \propto \exp(\eta G_j(t))$, where $\eta$ is a hyperparameter and  $G_j(t) = \sum_{t'=1}^t \mathrm{score}_j(\bm{x}_j(t'))$ is the total score of the action $j$ up to the time $t$.

The original GP-Hedge algorithm proposed in \cite{hoffman2011portfolio} was defined to solve a maximization problem, so the reward of each acquisition function is equal to the sum of the previous posterior means, i.e., $G_j(t) = \sum_{t'=1}^t \mu_j(\bm{x}_j(t'))$. Since we define our tasks as minimization problems, we multiply each score by $-1$ before adding it, i.e., $G_j(t) = - \sum_{t'=1}^t \mu_j(\bm{x}_j(t'))$. The full algorithm is detailed in Algorithm \ref{GP-Hedge}.

\begin{algorithm}[h]
\caption{GP-Hedge}
  \begin{algorithmic}
    \STATE Select hyperparameter $\eta \in \mathbb{R}^+$
    \STATE Set $G_j(0) = 0$ for $j = 1,2,\ldots,J$
    \FOR{t = 1,2,\ldots}
      \STATE Nominate points from each acquisition function $h_j$:
      
      $\bm{x}_j(t) = \arg\max_{\bm{x}} h_j(\bm{x}).$
      \STATE Select a nominee $\bm{x}(t) = \bm{x}_j(t)$ with probability
      
      $p_j(t) = \frac{\exp(\eta G_j(t-1))}{\sum_{j'=1}^J \exp(\eta G_{j'}(t-1))}.$
      \STATE Compute $y(t)$ by evaluating the objective on point $\bm{x}(t)$.
      \STATE Augment the data $\mathcal{D}_t$ with the new pair $(\bm{x}(t), y(t))$.
      \STATE Update the surrogate GP model.
      \STATE Update the rewards $G_j(t) = G_j(t-1) - \mu(\bm{x}_j(t))$ from the updated GP posterior.
    \ENDFOR
 \end{algorithmic}
 \label{GP-Hedge}
 \end{algorithm}

\section{Proposed Methodology}
\label{sec_proposed}

\subsection{GP-Hedge limitations}

The original GP-Hedge algorithm relies on the cumulative performance of each acquisition function in the portfolio during previous iterations to favor the choice of a function over the others. However, for large enough horizons, the influence of the first iterations may not be relevant or even desirable. Thus, we want to reduce the importance of an iteration in the reward function as the former becomes more distant from the current iteration. Importantly, it is emphasized in \cite{shahriari2014entropy} that the reward function is critical for the GP-Hedge performance, which encourages our argument.

\begin{figure*}
    \centering
    \begin{subfigure}{.32\linewidth}
        \includegraphics[width=\linewidth]{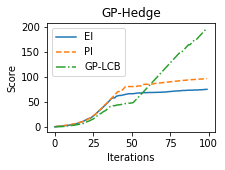}
    \end{subfigure}
    \begin{subfigure}{.32\linewidth}
        \includegraphics[width=\linewidth]{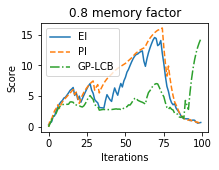}
    \end{subfigure}
    \begin{subfigure}{.32\linewidth}
        \includegraphics[width=\linewidth]{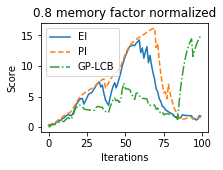}
    \end{subfigure}\\
    \begin{subfigure}{.32\linewidth}
        \includegraphics[width=\linewidth]{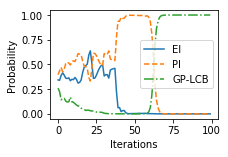}
    \end{subfigure}
    \begin{subfigure}{.32\linewidth}
        \includegraphics[width=\linewidth]{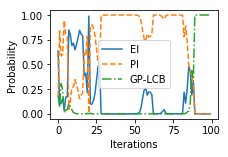}
    \end{subfigure}
    \begin{subfigure}{.32\linewidth}
        \includegraphics[width=\linewidth]{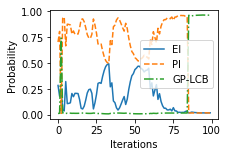}
    \end{subfigure}
    \caption{In the first row, it is presented the scores of each acquisition function along the $100$ iterations of a typical run for the Hartmann 6 benchmark function. The first column corresponds to the GP-Hedge, the second column corresponds to the GP-Hedge with a memory factor of 0.8, and, finally the third column corresponds to the memory factor of 0.8 using the normalization strategy, which consists in our approach. In the second row, the correspondent probabilities of each acquisition function be chosen is shown.}
    \label{fig:scores}
\end{figure*}

In Fig. \ref{fig:scores} it is illustrated an example of how the scores of each acquisition function evolve with the iterations. We consider a 3-function portfolio comprised by the PI, EI and GP-LCB functions and the well known Hartmann 6 benchmark. The corresponding probabilities of being chosen are also presented. It is possible to note that for the GP-Hedge the GP-LCB function finishes with a large score lead over the other 2 functions. In comparison with the versions with a memory factor, we note that this lead is smaller and can still be lost. Both score graphs for the versions with memory factors are very similar. However, the probability graphs are different, since in the normalized version it is more uncommon to obtain situations of equal probabilities, which would result in a completely random choice between the 3 available functions. Both the memory factor and normalization mechanisms will be detailed in the next sections.

\subsection{Memory Factor}

In order to tackle the presented GP-Hedge issues, we propose to change the reward function update by including a memory factor. The main goal of the memory factor is to decrease the influence of previous iterations in the reward evaluation as new iterations are completed, while still considering past experiences. Thus, we aim to avoid that discrepant values in the first few iterations of being determinant in the acquisition function selection during all the later iterations. Our approach also enables more probability for acquisition functions that are better in the recent past, according to the memory factor value, enabling ``recovery'' from far past mistakes.

More specifically, we propose to change the reward update computation as follows:
\begin{equation}
\label{reward_mem_fac}
G_j(t) = m G_j(t-1) - \mu(\bm{x}_j(t)),
\end{equation}
where $0 \leq m \leq 1$ is the memory factor hyperparameter, $\mu(\bm{x}_j(t))$ is the GP posterior mean for the input $\bm{x}_j(t)$ suggested by acquisition function $h_j$ at iteration $t$. Typical values for the memory factor hyperparameter are between $0.7$ and $1$, where the latter recovers the original GP-Hedge.

Eq. \eqref{reward_mem_fac} imposes a decrease in relevance to past rewards. For $m < 1$ we achieve two behaviors that are difficult to observe in the original GP-Hedge: (\textit{i}) initially bad acquisition functions may receive some probability in the later iterations if they improve along the optimization steps; (\textit{ii}) acquisition functions that go very well in the beginning may lose the preference if they are not able to keep the good performance. Those behaviors can be observed in Fig. \ref{fig:scores}, where we can note, for instance, that the EI function is able to recover some probability after the initial iterations in the variants that include the memory factor.

\subsection{Rewards Normalization}
\label{sec_rewards_normalization}

The use of the previously presented memory factor may cause the rewards of the portfolio to be very close at some iterations, as illustrated in Fig. \ref{fig:scores}. In such scenario, all the acquisition functions have about the same probability of being chosen. In the extreme case where all the functions have equal probability we would get an undesired completely random portfolio behavior.

In order to solve this problem, we propose to normalize the reward functions values before computing the choice probabilities as follows:
\begin{align}
\label{reward_mem_fac_norm}    
r_j(t) & = \frac{G_j(t) - r_{\text{max}}(t)}{r_{\text{max}}(t) - r_{\text{min}}(t)}, \\
\nonumber
\text{where } r_{\text{min}}(t) & = \min_j(G_j(t)), \\
\nonumber
r_{\text{max}}(t) & = \max_j(G_j(t)).
\end{align}
In the former expressions, the term $r_j(t)$ indicates the normalized reward for the acquisition function $j$ after the iteration $t$. Those values, computed for each acquisition function $h_j|_{j=1}^J$, are then used to obtain the probabilities of each acquisition function being chosen as follows:
\begin{equation}
\label{prob_norm} 
p_j(t) = \frac{\exp(\eta r_j(t-1)}{\sum_{j'=1}^J \exp(\eta r_{j'}(t-1))}.
\end{equation}
Note that the normalized $r_j(t)$ values are considered only to compute Eq. \eqref{prob_norm}. The original values of $G_j(t)$ are not overwritten.

The proposed normalization step constrains the terms $r_j(t)$ to be between $-1$ and $0$, with the highest value always being $0$ and the lowest always being $-1$. To solve a possible division by $0$, if the highest and lowest rewards are equal, all probabilities are set equally.

\subsection{The No-PASt-BO Algorithm}

The proposed changes in the rewards function of the base GP-Hedge approach result in the proposal of the No-PASt-BO (Normalized Portfolio Allocation Strategy for Bayesian Optimization) algorithm, which is detailed in Algorithm \ref{nopast_algorithm}. Note that the proposed method presents the same computational cost of the original GP-Hedge.

\begin{algorithm}[H]
\caption{No-PASt-BO}
  \begin{algorithmic}
    \STATE Select hyperparameter $\eta \in \mathbb{R}^+$
    \STATE Select hyperparameter $m \in [0,1]$
    \STATE Set $G_j(0) = 0$ for $j = 1,2,\ldots,J$
    \FOR{t = 1,2,\ldots}
      \STATE Nominate points from each acquisition function $h_j$:
      
      $\bm{x}_j(t) = \arg\max_{\bm{x}} h_j(\bm{x})$
      \STATE Compute $r_{\text{min}}(t-1) = \min_j(G_j(t-1))$
      \STATE Compute $r_{\text{max}}(t-1) = \max_j(G_j(t-1))$
      \STATE Compute the normalized rewards:
      
      $r_j(t-1) = \frac{G_j(t-1) - r_{\text{max}}(t-1)}{r_{\text{max}}(t-1) - r_{\text{min}}(t-1)}$
      \STATE Select a nominee $\bm{x}(t) = \bm{x}_j(t)$ with probability
      
      $p_j(t) = \frac{\exp(\eta r_j(t-1)}{\sum_{j'=1}^J \exp(\eta r_{j'}(t-1))}$.
       \STATE Compute $y(t)$ by evaluating the objective on point $\bm{x}(t)$.
      \STATE Augment the data $\mathcal{D}_t$ with the new pair $(\bm{x}(t), y(t))$.
      \STATE Update the surrogate GP model.
      \STATE Update the rewards $G_j(t) = m G_j(t-1) - \mu(\bm{x}_j(t))$ from the updated GP posterior.
    \ENDFOR
 \end{algorithmic}
 \label{nopast_algorithm}
 \end{algorithm}
 
\section{Experiments}
\label{sec_experiments} 
 
In order to evaluate the proposed No-PASt-BO performance a battery of experiments were made in synthetic benchmark function optimization. We also considered the real world application of BO in the task of optimizing the hyperparameters of machine learning models. For each test 25 runs were executed and the mean logarithmic error is reported along with the correspondent obtained confidence interval, i.e., the standard deviation scaled by the square root of the number of runs. 

As baselines, we also evaluate a random portfolio (RP) approach, which chooses randomly among a set of predefined acquisition functions, the original GP-Hedge and standard BO with a single acquisition function. For all of the experiments, we have used the GpyOpt package, a general BO framework introduced in \cite{gpyopt2016}. After preliminary experiments, the hyperparameter $\eta$ was set to $4$ for all of the No-PASt-BO experiments, while for the GP-Hedge the $\eta$ was defined using the strategy suggested in \cite{hoffman2011portfolio}.

\subsection{Memory Factor Sensibility}

Before the comparison results, we wish to explore the impact of the memory factor $m$ in the No-PASt-BO method. Thus, we perform an initial experiment with $7$ different values for the memory factor: $\{0.7, 0.75, 0.8, 0.85, 0.9, 0.95, 1\}$. Note that when the memory factor is equal to $1$, it means that previous executions do not lose importance. With the exception of the normalizing step (see Section \ref{sec_rewards_normalization}), the latter is similar to the GP-Hedge. As the memory factor decreases, the importance of previous evaluations also decreases.

\begin{figure}
    \centering
    \begin{subfigure}{.98\linewidth}
        \includegraphics[width=\linewidth]{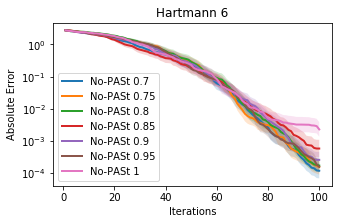}
    \end{subfigure}
    \begin{subfigure}{.48\linewidth}
        \includegraphics[width=\linewidth]{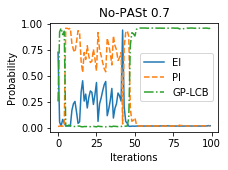}
    \end{subfigure}
    \begin{subfigure}{.48\linewidth}
        \includegraphics[width=\linewidth]{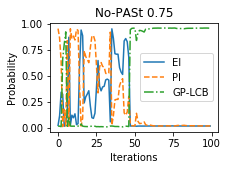}
    \end{subfigure}
    \begin{subfigure}{.48\linewidth}
        \includegraphics[width=\linewidth]{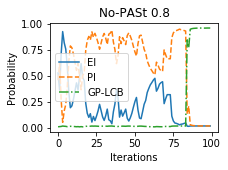}
    \end{subfigure}
    \begin{subfigure}{.48\linewidth}
        \includegraphics[width=\linewidth]{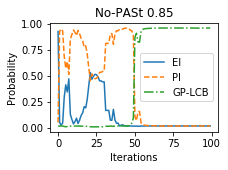}
    \end{subfigure}
    \begin{subfigure}{.48\linewidth}
        \includegraphics[width=\linewidth]{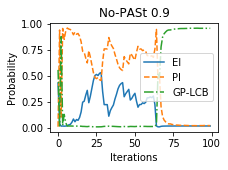}
    \end{subfigure}
    \begin{subfigure}{.48\linewidth}
        \includegraphics[width=\linewidth]{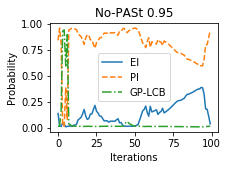}
    \end{subfigure}
    \begin{subfigure}{.48\linewidth}
        \includegraphics[width=\linewidth]{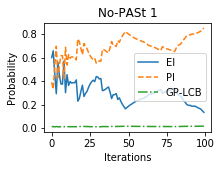}
    \end{subfigure}
    \caption{The effect of different memory factors in the NO-PASt-BO execution. As the memory factor increases, it takes more time to select an acquisition function as the one with higher probability.}
    \label{memory_factor_effect}
\end{figure}

The impact of different the memory factor values is presented in Fig. \ref{memory_factor_effect} for the Hartmann 6 benchmark function. It is possible to verify how the rewards change over the iterations, and how it affects the probability of the correspondent acquisition function being chosen. Higher values of memory factor make it more difficult to select an acquisition function as the most probable one. Moreover, lower values for the memory factor imply in more quickly varying probabilities, which may improve the diversity in the acquisition function selection step.

\subsection{Synthetic Benchmark Functions}

In this section $3$ standard benchmark functions were used: Branin, Hartmann 3 and Hartmann 6\footnote{Functions definitions available at \url{https://www.sfu.ca/~ssurjano/optimization.html}.}. Their domains are respectively 2, 3 and 6 dimensional. These functions were the same chosen in the original GP-Hedge paper \cite{hoffman2011portfolio}.

\begin{figure}
    \centering
    \begin{subfigure}{\linewidth}
        \includegraphics[width=0.98\linewidth]{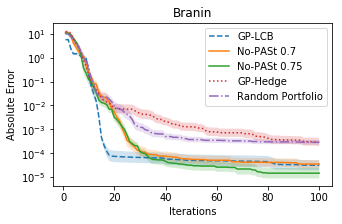}
    \end{subfigure}
    \begin{subfigure}{\linewidth}
        \includegraphics[width=0.98\linewidth]{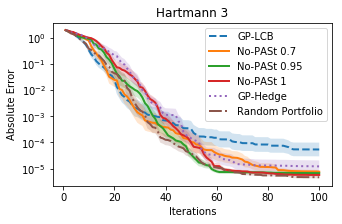}
    \end{subfigure}
    \begin{subfigure}{\linewidth}
        \includegraphics[width=0.98\linewidth]{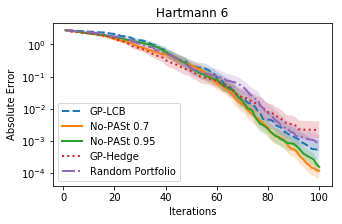}
    \end{subfigure}
    \caption{Results of the minimization of the synthetic benchmark using 3 acquisition functions. Note that only the 2 best choices for the NO-PASt-BO memory factor are shown. Also, only the best non-portfolio acquisition function is presented.}
    \label{fig:3func}
\end{figure}

\begin{figure}
    \centering
    \begin{subfigure}{\linewidth}
        \includegraphics[width=0.98\linewidth]{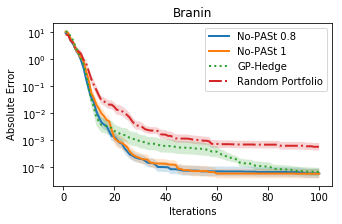}
        \label{branin_3f}
    \end{subfigure}
    \begin{subfigure}{\linewidth}
        \includegraphics[width=0.98\linewidth]{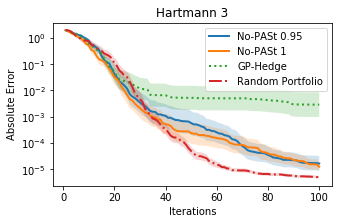}
        \label{hartmann3d_3f}
    \end{subfigure}
    \begin{subfigure}{\linewidth}
        \includegraphics[width=0.98\linewidth]{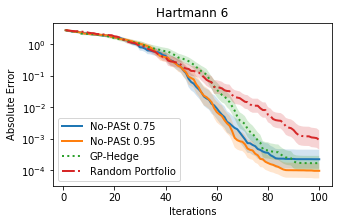}
        \label{hartmann6d_3f}
    \end{subfigure}
    \caption{Results of the minimization of the synthetic benchmark using 9 acquisition functions. Note that only the 2 best choices for the NO-PASt-BO memory factor are shown. Also, only the best non-portfolio acquisition function is presented.}
    \label{fig:9func}
\end{figure}

In all experiments, the portfolio-based strategies used the same set of acquisition functions: PI, EI and GP-LCB. We have set $\xi = 0.01$ for PI and EI, and $\delta = 0.1$ and $\nu = 0.2$ for GP-LCB. Furthermore, we evaluate the proposed No-PASt-BO with 7 different memory factors. For each of the experiments, 5 initial points were selected using the Latin Hypercube Sampling (LHS) approach \cite{mckay1979comparison}.

The results of these experiments can be found in Fig. \ref{fig:3func}. For all of scenarios, a portfolio strategy obtained the best performance. A version of the No-PASt-BO achieved the best result in two of the three case, and was very close to the best version in the case it loses. Also, the No-PASt-BO with memory factor of $0.7$ outperforms GP-Hedge in all of the executed experiments.

A second battery of tests was run for the same synthetic functions, but using 9 acquisition functions in the portfolio variants. These functions come from the same set of PI, EI and GP-LCB adding $\xi = 0.1$, $\xi = 1.0$,  $\nu = 0.1$ and $\nu = 1$ to the already studied values. The results are show in Fig. \ref{fig:9func}. Overall, the Random Portfolio performance was worse than the previous experiment with only 3 functions, with the remarkable exception of the Hartmann 3 experiment, due to the fact that as more functions were inserted in the portfolios, some of them are worse to the task at hand. In the same way, we can see that GP-Hedge got more competitive in general, but the No-PAST-BO still reaches lower optimized values faster.

\subsection{Real World Problems}

In order to evaluate the No-PASt-BO and compare it with the available baselines in a real world problem, we first consider a regression setting with the standard Boston Housing dataset\footnote{Available at \url{https://www.cs.toronto.edu/~delve/data/boston/bostonDetail.html}.}. The Support Vector Regression (SVR) model \cite{drucker1997support} is used to predict houses prices from the available attributes. The BO strategies have the task of optimizing the 3 SVR hyperparameters (gamma, C and epsilon), as implemented in the scikit learn toolbox \cite{scikit-learn}.

\begin{figure}
    \centering
    \begin{subfigure}{\linewidth}
        \includegraphics[width=0.98\linewidth]{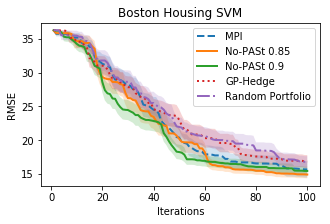}
    \end{subfigure}
    \begin{subfigure}{\linewidth}
        \includegraphics[width=0.98\linewidth]{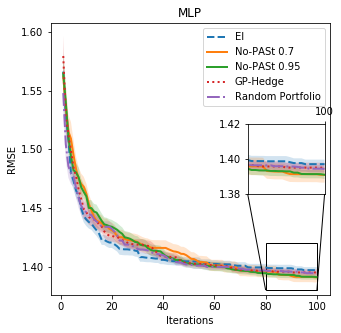}
    \end{subfigure}
    \caption{Results for automatic tuning of SVM and MLP hyperparameters. For both cases a No-PASt-BO version has achieved the best performance.}
    \label{fig:real}
\end{figure}

In this experiment we have used a $10$-fold cross-validation procedure, where the objective of each BO step is to minimize the average root mean square error (RMSE). For each experiment 100 optimization iterations were performed and each experiment was repeated 25 times. The obtained average RMSE values and their confidence intervals are shown in Fig. \ref{fig:real}. We can see that the NO-PASt-BO variants achieved the best results both in terms of best final solution and faster lower values. Moreover, the GP-Hedge performed comparable to the the random portfolio strategy.

We tackled a second real world problem to compare the evaluated BO methods in a regression setting which consists in predicting the average gearbox high-speed shaft temperature in a Wind Turbine Generator (WTG) from a given set of measures. Following the same methodology presented in \cite{bangalore2017,bracis2019}, the input variables were: average active power, average rotor speed, average nacelle (the turbine outer casing) temperature and average outdoor temperature. A limit filter was applied to the available measured data, removing values that are physically absurd. After that, we applied a clustering filter to remove some outliers from the dataset and, finally, we applied a continuity filter in order to remove isolated points. Details of these preprocessing steps can be seen in \cite{bangalore2017,bracis2019}.

We considered a nonlinear model comprised of a Multilayer Perceptron (MLP) network with a single hidden layer and hyperbolic tangent hidden activations. The model was trained via stochastic gradient descent optimizer. The task of the BO was to minimize the RMSE of the predicted temperature in a hold-out validation set, varying over the scikit-learn MLPRegressor hyperparameters \cite{scikit-learn}: neurons in the hidden layer, learning rate, alpha (weight decay hyperparameter) and momentum. The data used in this experiment was obtained from a WTG located in Brazil. Results are presented in Fig. \ref{fig:real}. Although all the methods obtained somehow comparable results, the detailed ispection indicate that the NO-PASt-BO variants obtained the best RMSE values at the end.

\section{Conclusions and Further Work}
\label{sec_conc}

In this work we have proposed No-PASt-BO, a new methodology for performing black-box Bayesian optimization of unknown functions. Our approach builds on the popular GP-Hedge method \cite{hoffman2011portfolio} to create an adaptive portfolio of acquisition functions. We aimed to tackle some of the GP-Hedge's limitations, such as the sometimes undesirable high influence of far past evaluations, by incorporating a limited memory and a normalization mechanisms.

We evaluate the proposed approach on both synthetic and real-world optimization tasks, where No-PASt-BO obtained competitive performance with respect to the other evaluated strategies. Importantly, our approach always outperforms GP-Hedge, while maintaining its simplicity and general applicability. The latter features may enable No-PASt-BO to become the default off-the-shelf method for portfolio-based BO\footnote{The No-PASt-BO source code is available at \url{https://github.com/thiago-vasconcelos/no-past-bo}.}.

Further investigations shall verify how to incorporate non-myopic concepts to our portfolio BO methodology, such as the ones explored in \cite{gonzalez2016glasses}. That would enable more clever decision making when we have a known limited budget in terms of number of objective evaluations.

Another interesting subject for investigation is the task of function optimization with constraints known \textit{a priori}. Strategies such as the ones presented in \cite{gardner2014bayesian} may turn our portfolio framework even more applicable in real world problems.

\bibliography{references}

\end{document}